\documentclass[journal]{IEEEtran}
\usepackage{graphicx}
\usepackage{caption}
\usepackage{url}
\usepackage{booktabs}
\usepackage{adjustbox} % 引入adjustbox宏包
\usepackage{hyperref}
\ifCLASSINFOpdf
\else
\fi
\begin{document}
\title{Generalizable and Explainable Deep Learning for Medical Image Computing: An Overview}

\author{Ahmad Chaddad, Yan Hu, Yihang Wu, Binbin Wen, Reem Kateb% <-this % stops a space
\thanks{A.Chaddad, Y.Hu, Y.Wu, B.Wen are with the Artificial Intelligence for Personalized Medicine, School of Artificial Intelligence, Guilin University of Electronic Technology, Guilin 541004, China.}% <-this % stops a space
\thanks{A.Chaddad is with the Laboratory for Imagery Vision and Artificial Intelligence, École de Technologie Supérieure (ETS), Montréal, QC H3C 1K3, Canada.
}% <-this % stops a space
\thanks{R. Kateb is with the College of Computer Science and Engineering, Taibah University, Madinah, 42353, Saudi Arabia, and College of Computer Science and Engineering, Jeddah University, Jeddah, 23445, Saudi Arabia.}
\thanks{All authors are equally contributed.
}
}


% \usepackage{cite}

\markboth{Journal of \LaTeX\ Class Files,~Vol.~14, No.~8, August~2015}%
{Shell \MakeLowercase{\textit{et al.}}: Bare Demo of IEEEtran.cls for IEEE Journals}



\maketitle

\begin{abstract}
Objective. This paper presents an overview of generalizable and explainable artificial intelligence (XAI) in deep learning (DL) for medical imaging, aimed at addressing the urgent need for transparency and explainability in clinical applications.

Methodology. We propose to use four CNNs in three medical datasets (brain tumor, skin cancer, and chest x-ray) for medical image classification tasks. In addition, we perform paired t-tests to show the significance of the differences observed between different methods. Furthermore, we propose to combine ResNet50 with five common XAI techniques to obtain explainable results for model prediction, aiming at improving model transparency. We also involve a quantitative metric (confidence increase) to evaluate the usefulness of XAI techniques.

Key findings. The experimental results indicate that ResNet50 can achieve feasible accuracy and F1 score in all datasets (e.g., 86.31\% accuracy in skin cancer). Furthermore, the findings show that while certain XAI methods, such as XgradCAM, effectively highlight relevant abnormal regions in medical images, others, like EigenGradCAM, may perform less effectively in specific scenarios. In addition, XgradCAM indicates higher confidence increase (e.g., 0.12 in glioma tumor) compared to GradCAM++ (0.09) and LayerCAM (0.08).

Implications. Based on the experimental results and recent advancements, we outline future research directions to enhance the robustness and generalizability of DL models in the field of biomedical imaging.

\end{abstract}

\begin{IEEEkeywords}
eXplainable artificial intelligence; deep learning; generalizable model
\end{IEEEkeywords}

% \IEEEpeerreviewmaketitle

\section{Introduction}

\subsection{Why the field of deep learning needs explainable AI?}
Deep learning (DL) has seen a significant increase in popularity within the fields of artificial intelligence (AI) and machine learning (ML). It has demonstrated remarkable performance in areas such as computer vision, natural language processing, and gaming \cite{chaddad2023federated}. Although DL has greatly advanced medical image analysis, its adoption in clinical radiology is limited by concerns over model explainability and generalizability \cite{ellis2021explainable}. Since most deep models lack explainability, they are often treated as black boxes. In critical applications related to fairness, privacy, and safety, it is essential to understand the underlying mechanisms behind the predictions of deep models in order to trust them fully. Therefore, accurate predictions and human-intelligible explanations are necessary, particularly for users in interdisciplinary domains, which highlights the importance of developing explanation techniques for deep neural networks (DNNs) \cite{yuan2022explainability}. This has led to a growing demand for approaches to better understand those black boxes, known as interpretable deep learning or explainable artificial intelligence - XAI \cite{chaddad2023explainable}. For example, the analysis of medical images across disciplines such as dermatology, ophthalmology, and radiology has increasingly relied on DL, specifically DNNs. Technically, DNNs learn representations of statistical patterns and inherent structures from large datasets. In particular, deep convolutional neural networks (CNNs) are state-of-the-art models for image segmentation and classification and are suitable for abstracting highly complex spatial patterns from images \cite{nazir2023survey}. These models are trained in a supervised manner by repeatedly presenting data points (e.g., images) along with their corresponding labels (e.g., "cancer or non-cancer"). Throughout this learning process, the internal parameters (weights) of the CNN are iteratively adjusted by minimizing a loss function, which measures the deviation of the model predictions from the known labels. However, the application of DL in medical decision-making, traditionally relies on systems designed entirely by human experts, has raised concerns due to its ambiguous nature.

To address such concerns, XAI methodologies in medical imaging are increasingly explored to illustrate how DL algorithms reach their decisions. DL ambiguity faces particular challenges in fields where decision-making carries strong risks. Therefore, it is understandable that clinicians are concerned about the lack of transparency in DL. Regulatory frameworks like the European Union General Data Protection Regulation - GDPR, \cite{voigt2017eu} emphasize transparency. They ensure patients have the right to understand the processes behind decisions influencing them. For example, techniques like Gradient-weighted Class Activation Mapping (Grad-CAM) \cite{selvaraju2017grad}, and Local Interpretable Model-Agnostic Explanations (LIME) \cite{ribeiro2016should} have been integrated into medical imaging to enhance model explainability. Grad-CAM highlights image regions important to a model decision by generating heatmaps, while LIME explains predictions by approximating the model locally with simpler interpretable models. These methods help make AI systems more transparent for clinicians.

\subsection{Challenges and solutions of XAI techniques
}

We first explain the three terms used in the context of XAI: 1) explainability, 2) interpretability, and 3) transparency. Specifically, explainability focuses on clarifying the internal workings of complex models, while interpretability pertains to the ease of understanding these models, particularly in the case of simpler architectures. Transparency refers to the extent to which the model operations can be comprehended by users.

However, XAI techniques in medical imaging tasks face several challenges. Methods like Grad-CAM rely primarily on local explanations for specific inputs, often limited to interpreting the model decision on a single instance and failing to reflect the model overall performance and reliability [citation]. 

In this work, we consider five XAI methods (GradCAM++ \cite{chattopadhay2018grad}, EigenGradCAM \cite{muhammad2020eigen}, XGradCAM \cite{fu2020axiom},  AblationCAM \cite{ramaswamy2020ablation}, LayerCAM \cite{jiang2021layercam}) to integrate different local explanation techniques, providing a more comprehensive model understanding. Moreover, there is a lack of widely accepted quantitative metrics to evaluate the quality of XAI explanations \cite{chaddad2023survey}, as most assessments rely on qualitative human analysis. In this paper, we employ the Remove and Debias (ROAD) method \cite{rong2022consistent} to quantify and assess XAI techniques. The ROAD method evaluates the predictive confidence of a model for specific inputs by removing important features and observing changes in model performance. This variation in confidence reflects the impact of the removed features on the model decision-making process, providing a concrete quantitative standard for assessing the effectiveness and reliability of XAI techniques. A higher confidence level indicates superior performance of the corresponding XAI method.

Furthermore, there are several limitations in applying DL in medical settings. One key concern is uncertainty about how well the developed models can be applied to different datasets than the ones they were trained on. It is recommended to use independent (external) datasets to evaluate these models. For example, the model trained and tested on the same data may become too specialized and give overly optimistic performance estimates (“overfitting”). The limited generalizability of DL models may be due to differences in image or population features. Specifically, the DL models may not adapt well to variations or new conditions outside their training field \cite{bassi2024improving}. Understanding the causes of these limitations can help improve modeling strategies and establish standards for model benchmarking before clinical use, ensuring robustness and generalizability \cite{chekroud2024illusory}. 

\subsection{Research contributions
}

This study aims to fill research gaps by providing a comprehensive overview and implementation of popular XAI techniques in medical image classifications. It focuses on showing five most common XAI methods (GradCAM++ , EigenGradCAM, XGradCAM, AblationCAM, LayerCAM) to enhance the global explainability of XAI techniques and employs the ROAD method to quantitatively and systematically evaluate the reliability of these techniques. To evaluate the efficiency of these XAI methods, the time taken by each method to produce results is presented. Moreover, to evaluate model generalization performance, we conducted experiments using four CNN models: Residual Network 50 (ResNet50), Densely Connected Convolutional Network 121 (DenseNet121), Visual Geometry Group 16-layer network (VGG16), and Efficient Network version B0 (EfficientNet-B0) with three public datasets. We performed statistical analysis (paired t-tests), to compare the model performance. P-values were calculated to evaluate the significance of the pairwise differences, determining whether there were statistically significant variations in performance between the models. 

The contributions can be summarized as follows.
\begin{enumerate}
    \item We simulate five common XAI techniques using three public medical data sets for medical image classification tasks.
    \item We employ the ROAD method to quantitatively evaluate the five XAI techniques, offering a framework for assessing the explainability of each technique.
    \item We evaluate model generalization by testing the four CNN models with three public datasets, using paired t-tests to assess statistically significant performance differences.
    \item We highlight the challenges and advantages of common XAI techniques in classifying medical images, along with the future directions.
\end{enumerate}

The paper is organized as follows. Section 2 investigates the recent progress of XAI and points out the potential threats of those studies. Section 3 illustrates the definition of generalizability in deep learning and the approaches to achieve a generalizable model with XAI in medical imaging. Section 4 presents a case study using five common XAI techniques with three public medical datasets. Section 5 analyzes the performance and efficiency of XAI techniques and summarizes the limitations and future directions of DL and XAI. Finally, in Section 6, we provide a comprehensive analysis and our concluding remarks.

\begin{table}[]
    \centering
    \caption{Summary of the abbreviations used in this study.}
    \includegraphics[width=0.97\linewidth]{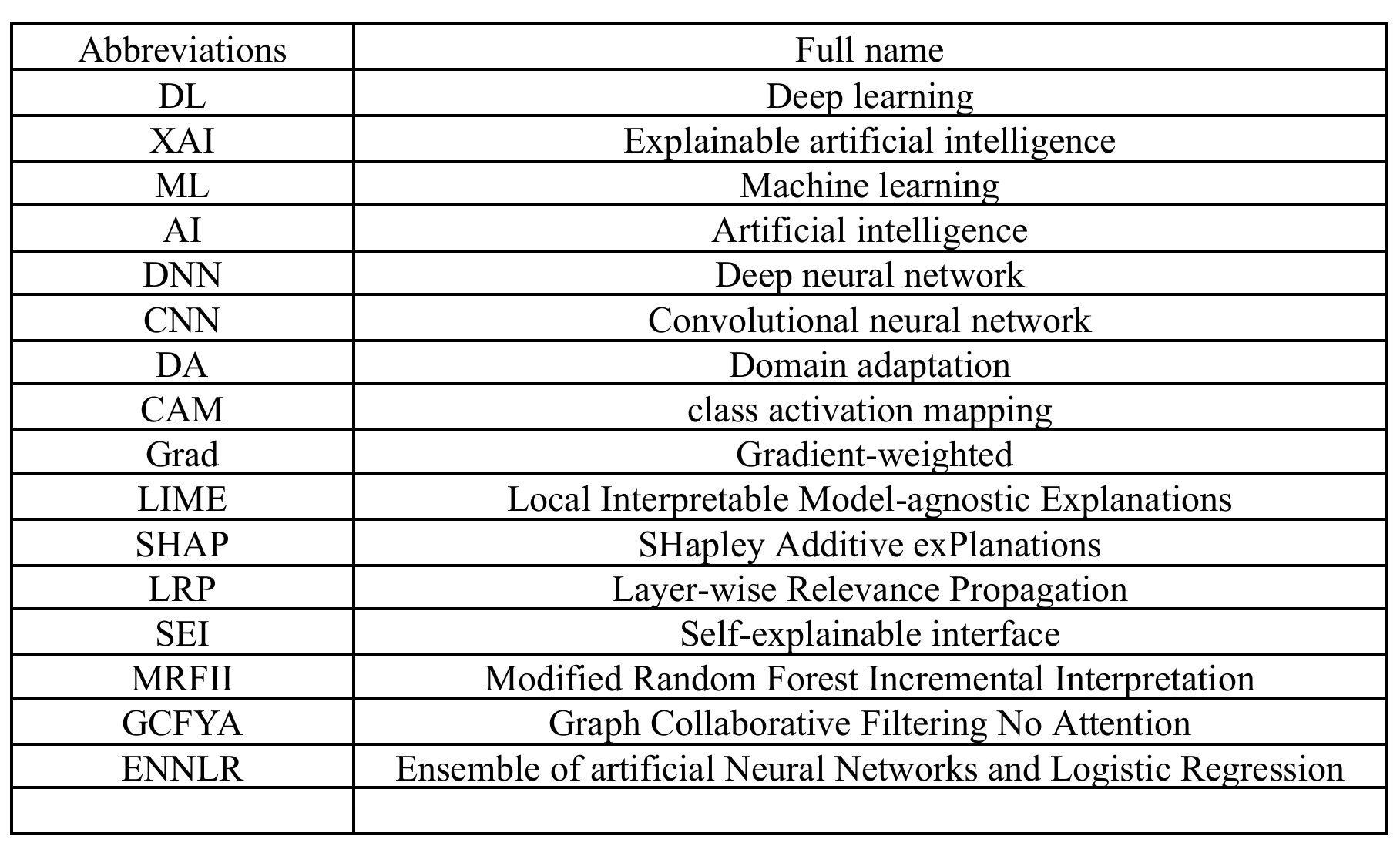}
    
    \label{tab:abbres}
\end{table}

\section{Related work}

Various XAI techniques have been developed to enhance the interpretability of AI models, especially in healthcare. These methods broadly fall into two categories: post-hoc explanation techniques and intrinsic interpretable models \cite{chaddad2023survey}. However, we focus on the most common techniques used recently, particularly post-hoc methods, which aim to explain black-box models like CNNs.

For example, GradCAM model proposes to use the gradients of the target class with respect to the feature maps of a convolutional layer \cite{selvaraju2017grad} . Furthermore, in \cite{lundberg2017unified}, they propose to explain the models prediction by computing the contribution of each feature to the prediction (SHAP). In addition, in 
\cite{ribeiro2016should}, they introduce a novel explainable methodology named LIME, which provides interpretable explanations by approximating it locally with a simple, understandable model. Furthermore, based on GradCAM \cite{chattopadhay2018grad}, they provide GradCAM++ to involve gradient information (e.g., weights) to enhance the interpretability of GradCAM. In addition, in \cite{muhammad2020eigen}, they explain a novel method called EigenGradCAM. It can use principal component analysis (PCA) on the spatial covariance of feature maps to generate more precise class activation maps for interpreting CNN predictions. In \cite{ramaswamy2020ablation}, they iteratively remove layers and observe the changes in class activation maps (AblationCAM), further providing a comprehensive interpretability result. Similarly, in \cite{fu2020axiom}, to provide robust class activation maps, they suggest using generalized gradients and a learn feature importance weighting mechanism to enhance the explainability of deep models (XgradCAM). Unlike the previous studies, the authors in \cite{jiang2021layercam} explore layer-wise explainability, proposing to aggregate feature maps from layers to build a comprehensive CAM (LayerCAM). In \cite{li2023multilayer}, they extend the GradCAM into multi-layer GradCAM to provide comprehensive interpretability. Generally, these XAI methods are the most commonly used in recent studies. Generally, these XAI methods have been used the most in recent studies. For example, Figure 1 illustrates the timeline of recent XAI techniques. The flags indicate top-cited (e.g., $>$ 500 citations) papers in the XAI field. Based on the timeline, we select GradCAM++, EigenGradCAM, AblationCAM, XgradCAM and LayerCAM as examples for comparison and evaluation in this study.

\begin{figure*}
    \centering
    \includegraphics[width=0.97\linewidth]{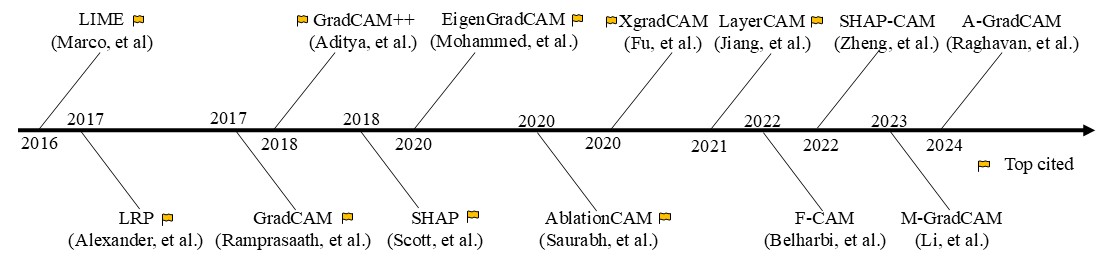}
    \caption{Timeline of recent XAI techniques. 
}
    \label{fig:timeline}
\end{figure*}

Despite the significant success of XAI models in numerous studies, several challenges remain that must be addressed. 1) Data bias: many studies rely on specific datasets, which may have problems with sample imbalance or insufficient data representation, limiting the generalizability of the results. 2) Experimental limitations: Many studies often focus on a single task or dataset, resulting in a limited amount of data and consequently restricting the range of application options. 3) XAI technical limitations: existing explanatory techniques, such as Grad-CAM and SHAP, provide model explainability, but may still lack in consistency for interpreting the data (e.g., generates different explanations).

In this study, we focus on further promoting the application of XAI technique in medical image classification, especially in improving the explainability of the model and enhancing clinician trust. By introducing a variety of XAI techniques, we explore how to better explain the decision-making process of DL models.

\begin{figure}
    \centering
    \includegraphics[width=0.97\linewidth]{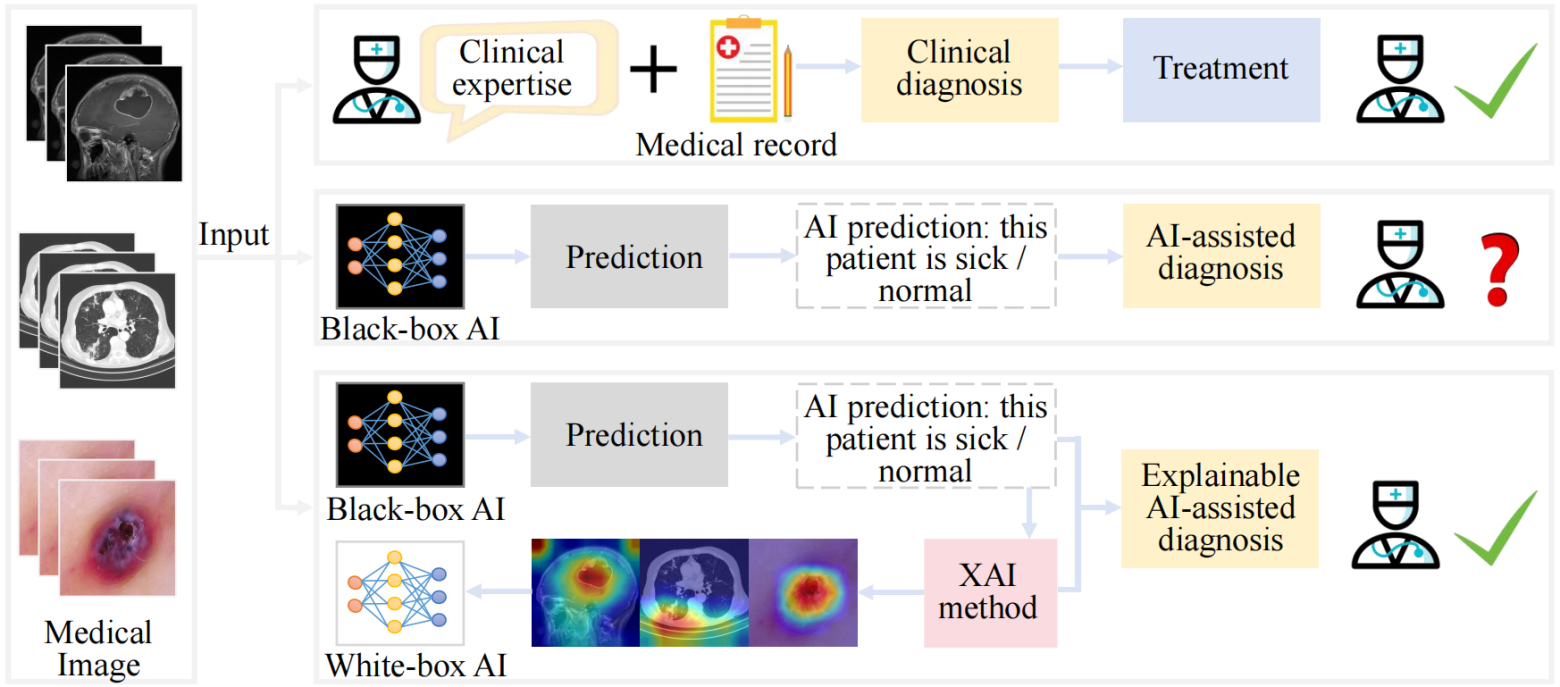}
    \caption{Example of XAI impact in clinical decision. Input images to three diagnostic scenarios: (First row) Clinicians use medical images and records for diagnosis, (Second row)  AI-assisted diagnosis handles complex data but lacks transparency, making trust difficult. (Third row) Explainable AI (XAI) enhances trust by providing interpretable predictions, combining AI's data processing with clinician expertise for better decisions.
}
    \label{fig:pipeline}
\end{figure}

\begin{table*}[h]\footnotesize

\caption{Summary of Explainable AI Models}
\begin{adjustbox}{max width=\textwidth} % 使用adjustbox调整表格宽度
\begin{tabular}{|l|c|c|c|c|c|l|}
\hline
\textbf{Models} & \textbf{DTS} & \textbf{C} & \textbf{I} & \textbf{ES} & \textbf{Topic} & \textbf{Principle} \\ \hline
LRP \cite{bach2015pixel,mandloi2024explainable} & $\leftrightarrow$ & $\uparrow$ & $\downarrow$ & $\uparrow$ & Brain tumor & Back propagates relevance through layers to explain model predictions \\ \hline
Grad-Cam \cite{selvaraju2017grad,el2024exhyptnet,nafisah2024tuberculosis} & $\downarrow$ & $\downarrow$ & $\uparrow$ & $\leftrightarrow$ & Hypertension Tuberculosis & Uses gradients for class-specific heatmaps \\ \hline
LIME \cite{ribeiro2016should,mahim2024unlocking} & $\uparrow$ & $\leftrightarrow$ & $\leftrightarrow$ & $\downarrow$ & Alzheimer's disease & Generates local surrogate models to explain predictions \\ \hline
SHAP \cite{lundberg2017unified,wani2024deepxplainer} & $\uparrow$ & $\uparrow$ & $\uparrow$ & $\uparrow$ & Alzheimer's \& Lung cancer & Computes Shapley values for feature attribution \\ \hline
Self-explainable interface \cite{dharmarathne2024novel} & $\uparrow$ & $\uparrow$ & $\uparrow$ & $\uparrow$ & diabetes & Integrates existing methods into an intuitive, user-friendly explanatory interface \\ \hline
MRFII \cite{chen2024deep} & $\uparrow$ & $\uparrow$ & $\leftrightarrow$ & $\uparrow$ & diabetes & Integrates decision rules with NN outputs for explanations \\ \hline
ENNLR \cite{shakhovska2024novel} & $\uparrow$ & $\uparrow$ & $\leftrightarrow$ & $\uparrow$ & Knee-disease & Combines NN and logistic regression for enhanced explainability \\ \hline
GCFYA \cite{huang2024interpretable} & $\uparrow$ & $\uparrow$ & $\leftrightarrow$ & $\uparrow$ & Chronic disease & Combines GNN with attention for enhanced explainability \\ \hline
GCFNA \cite{huang2024interpretable} & $\uparrow$ & $\leftrightarrow$ & $\leftrightarrow$ & $\leftrightarrow$ & Chronic disease & Uses GNN for node and edge explanations \\ \hline
\end{tabular}
\end{adjustbox}
{DTS: Data Type Applicability, C: Complexity, I: Intuitiveness, ES: Explanation Stability, $\uparrow$: high, $\downarrow$: low, $\leftrightarrow$: medium, Grad-Cam: Gradient-weighted Class activation mapping, LRP: Layer-wise Relevance Propagation, LIME: Local Interpretable Model-agnostic Explanations: SHAP SHapley Additive exPlanations, GCFYA: Graph Collaborative Filtering with Attention, GCFNA: Graph Collaborative Filtering No Attention, MRFII: Modified Random Forest Incremental Interpretation, ENNLR: Ensemble of artificial Neural Networks and Logistic Regression, NN: Neural networks, GNN: Graph neural networks.}
\label{tab:summary}
\end{table*}

\section{ReGeneralizable model with XAI}

\subsection{Explanation of generalizability in deep learning
}

The explanation techniques of DL models study the relationships behind the predictions of deep models. Many techniques are proposed to explain the deep models for image and text data \cite{yuan2022explainability}. These methods can provide input-dependent explanations, such as studying the important scores for input features, or a high-level understanding of the general behaviors of deep models. For example, by studying the gradients or weights \cite{yang2019xfake}, we can analyze the sensitivity between the input features and the predictions. Existing approaches \cite{du2018towards} map hidden feature maps to the input space and highlight the dominant input features. In addition, by occluding different input features, we can observe and monitor the change of predictions to identify the dominant features \cite{yuan2020interpreting}. Meanwhile, many studies \cite{simonyan2013deep} focus on providing input-independent explanations, such as studying the input patterns that maximize the predicted score of a certain class.

\subsection{Importance of explainability in medical image analysis
}

As mentioned previously, the problem of adapting DL technique lies in the ambiguity of the models as they are treated as “black boxes”. Therefore, it is difficult for clinicians to understand how the algorithms extracted the results. This issue arises simply because the mechanisms and reasoning processes are often hidden from the users; hence, there is no way to explain how it happens when it fails. Moreover, even a good AI typically produces many more false positives than true positives. Consequently, this issue reduces clinician’s trust and inhibits clinical adoption of AI that could save lives. Therefore, opening the black box to add explainability to the technology is important for physicians to build trust, thereby improving the adoption of DL techniques in healthcare. With explainable models, a clinician can make a more informed decision whether to accept or reject the result. As an example of XAI, Figure \ref{fig:pipeline} illustrates three diagnostic approaches: 1) Clinicians rely on solely medical images and patient medical records, using consultations and examinations to make clinical diagnoses and determine treatment plans. This method is transparent and clear, but clinicians may misdiagnose rare cases due to limited experience; 2) AI-assisted diagnosis, where AI models predict whether a patient is sick based on medical images. AI can handle large volumes of data and complex patterns, aiding in the detection of details that clinicians might overlook, particularly excelling in rare cases. However, the black-box nature of AI models makes it challenge for clinicians to understand and trust their predictions; 3) Explainable AI-assisted diagnosis, where AI models provide predictions and use explainability methods (e.g., visualizing the model focus areas in medical images) to interpret the results. This enhances clinicians trust in AI outcomes, combining AI powerful data processing capabilities with clinicians expertise, thereby better supporting clinical decision-making.

\subsection{Improve generalizability in AI models
}

There are many techniques to enhance the generalizability of deep models in medical image analysis. For example, federated learning (FL), as a decentralized learning framework that aims to learn a robust global model with multiple clients, can solve the overfitting issue in training deep models \cite{chaddad2023federated}. Similarly, domain adaptation (DA), as a feature alignment method, can greatly decrease the feature distribution discrepancies between different datasets, thereby improving the generalization ability of deep models \cite{kollias2024domain}. Likewise, introducing more features such as text features can also boost the generalization performance of deep models, as evidenced by contrastive language image pre-training methodology \cite{zhang2024vision}. Despite the approaches in previous studies, there are many other strategies such as self-supervised learning \cite{zhang2024self}, semi-supervised learning \cite{han2024deep} to further enhance the generalizability in medical image analysis. However, further research is needed to develop more techniques to generalize AI models, especially for clinical topics.

\subsection{Explainability algorithms in deep learning
}

In the domain of DL, explainability refers to the extent to which the internal decision-making processes of models and the impact of important features on outputs are comprehensible \cite{dhar2023challenges}. However, DL models are characterized by elaborate architectures and a large number of parameters, representing their decision-making processes opaque and difficult to interpret. This opacity presents a considerable challenge in advancing the explainability models \cite{hosain2024explainable}. With the widespread deployment of these models across various applications like image recognition, natural language processing, and speech recognition, there is an escalating imperative for transparency in understanding their decision mechanisms and articulating explanations of their outputs clearly.

These techniques for achieving model explainability range from feature visualization to local and global explanation methods. Through convolution kernel visualization, we can explore the features that each convolution kernel in a CNN focuses on. For example, heatmaps generated by class activation mapping (CAM) and Grad-CAM show the key areas in the decision-making process, providing an intuitive visual display for understanding how the model responds to different inputs. Local explanation methods, such as LIME and SHAP, approximate and explain the behavior of complex models by training simple models on the local behavior of the model \cite{chaddad2023survey}. Global explanation methods, including partial dependence plot (PDP) and principal component analysis (PCA), provide insights into the overall behavior of the model. In addition, model simplification and gradient-related methods, such as DeepLIFT, further evaluate the importance of features by analyzing the contribution of input features to model predictions \cite{li2021deep}. Together, these diverse methods promote a deeper understanding of the decision-making process of DL models.

In addition, probability theory, Bayesian theory, and other statistical methods are used to explain DL model decisions, particularly in uncertainty assessment, classification, and prediction tasks. Lambert et al. \cite{lambert2024trustworthy} review uncertainty quantification methods for DL models in medical image classification and segmentation (e.g., brain tumors, skin cancer, COVID-19), and propose a structural uncertainty framework to enhance model credibility and interpretability in clinical practice. Gao et al. \cite{gao2023bayeseg} introduces a Bayesian-based medical image segmentation framework (BayeSeg) that improves model generalization and interpretability by jointly modeling image and label statistics, excelling in prostate and heart segmentation, especially across modalities. While probability and Bayesian methods focus on quantifying uncertainty to improve model generalization and credibility, XAI methods aim to explain the decision-making process, making models more transparent. Together, they provide a more complete system for explaining and assessing uncertainty in clinical DL models. Table \ref{tab:abbres} lists all the abbreviations that used in this study.

Table \ref{tab:summary} presents a comparative analysis of XAI methods based on criteria such as data applicability, computational complexity, intuitiveness, and explanation stability. Methods like LIME, SHAP, and Modified Random Forest Incremental Interpretation (MRFII) exhibit wide applicability across various data types. High-complexity methods require more computational resources and advanced comprehension, whereas methods with high explainability afford more transparent and accessible explanations. Furthermore, methods characterized by high explanation stability provide consistent interpretative outcomes, ensuring reliability across different examples. This comparative assessment illustrates the relative strengths and limitations of each XAI method, thereby assisting clinicians/scientists in selecting the most suitable explainable approach for their applications.

\subsection{Approaches for integration in medical image computing
}

In brain image research, developing brain-computer interface (BCI) is important to understand brain states. Therefore, advanced neuroimaging techniques such as EEG, fMRI, and MEG are used to decode brain states. EEG, in particular, is widely used due to its non-invasiveness and cost-effectiveness, providing real-time insights into brain activity. Machine learning approaches are useful in analyzing EEG signals and extracting important features \cite{zhang2022explainable}. Despite many proposals for deep learning methods to analyze EEG data, it remains challenging to identify generalizable patterns across large datasets. Multiple random fragment search-based multilayer RNN, was proposed to solve this issue \cite{zhang2022explainable}.

In another example \cite{chen2022adversarial}, the authors introduce a method that uses adversarial learning and graph attention networks to identify autism spectrum disorder from brain imaging data. The method focuses on enhancing the model robustness against dataset variabilities to improve its generalization capabilities. Tan et al. \cite{tan2022fourier} present a novel Fourier domain robust denoising decomposition and adaptive patch MRI reconstruction method to address MRI noise and improve reconstruction with limited undersampled data. Another study introduces a method called gradient-matching federated DA for classifying brain images \cite{zeng2022gradient}. In \cite{chaddad2023enhancing}, the authors introduce the GradCAM technique to further enhance the explainability of domain adaptation models. Experimental results using COVID-19, skin cancer datasets demonstrate that the use of XAI can provide reasonable meaning for medical image classification. Similarly, in \cite{wu2024facmic}, the authors introduce a feature attention module to identify the most relevant features while using DA to mitigate the feature discrepancies between each client for classifying brain tumor and skin cancer images in FL context. Extensive experiments indicate the proposed method can provide a more generalized model compared with state-of-the-art methodologies. Their study highlights the potential of using FL with DA to improve the generalization ability of deep models.

\section{Case studies and experimental results}

Despite the developments of XAI methods, however, the key concept remains the same: to visualize the model prediction with readable and correct explainability results. We provide a case study using the common XAI techniques (GradCAM++ \cite{chattopadhay2018grad}, EigenGradCAM \cite{muhammad2020eigen}, XGradCAM \cite{fu2020axiom}, AblationCAM \cite{ramaswamy2020ablation} and LayerCAM \cite{jiang2021layercam}) and three public medical datasets (brain tumor, skin cancer and Chest X-ray) to demonstrate the usefulness of the XAI techniques. The following paragraph will elaborate on the implementation details and experimental results.

\subsection{ Samples and datasets}

To provide a comprehensive evaluation of XAI models in medical imaging, we select three common medical classification datasets, covering three modalities (Magnetic Resonance Imaging (MRI), Dermoscopy and Computed Tomography (CT). Specifically, 1) Brain tumor dataset has four classes, namely glioma tumor, meningioma tumor, no tumor and pituitary tumor\footnote{\url{https://www.kaggle.com/datasets/masoudnickparvar/brain-tumor-mri-dataset}}. The training set has 2870 samples, while the test set has 394 samples; 2) Skin cancer dataset comprises 10015 dermatoscopic images classified into seven groups: actinic keratoses and intraepithelial carcinoma / Bowen's disease  (AKIEC), basal cell carcinoma (BCC), benign keratosis-like lesions (BKL), dermatofibroma (DF), melanoma (MEL), melanocytic nevi (NV), and vascular lesions (VASC) \cite{tschandl2018ham10000}. Our training dataset consists of 8512 samples, while the test dataset comprises 1503 samples. 3) The Chest X-Ray dataset is a chest x-ray medical dataset, which has a training set consisting of 5232 samples and a testing set consisting of 624 images\footnote{\url{https://www.kaggle.com/datasets/paultimothymooney/chest-xray-pneumonia}}. It has two classes, normal and pneumonia. For each dataset, we split the original training set into two subsets, i.e., a training set (80\%) and a validation set (20\%). We repeat the experiments three times and report the average value (AVG) and standard deviation (STD) in Table 3.

\subsection{Implementation details 
}

We used ResNet50 \cite{he2016deep} as the main network backbone. The DenseNet121 \cite{huang2017densely}, VGG16 \cite{simonyan2014very} and EfficientNet-B0 \cite{tan2019efficientnet} are also considered for comparison with stochastic gradient descent (SGD) optimizer, a learning rate of $1×10^{-3}$, a weight decay of $5×10^{-4}$, and a momentum of 0.9. The data are preprocessed using PyTorch functions including image resize (224×224) and normalization using z-score. No data augmentation techniques are considered in this study. We set batch size and training epoch as 32 and 30, respectively. Furthermore, we used Python3.8 with PyTorch 1.13.1. All experiments are based on the Windows 11 operating system, and features an Intel 13900KF CPU with 128 GB of RAM and an RTX 4090 GPU.

\subsection{Results
}

Table \ref{tab:2} reports the testing accuracy, precision, recall and macro F1 using the three datasets. For example, the ResNet50 model can achieve feasible accuracy (e.g., 83.92\% precision to classify four classes of brain tumor) on the medical images. Despite the remarkable performance of those classifier models, however, we didn't know based on which part in the images the model is considered the decision. For this reason, we applied the XAI techniques to interpret the model decision.

To quantify XAI in medical imaging, we followed the Remove and Debias (ROAD) method to compare the confidence increase of those XAI techniques. The basic idea of their proposed method is to remove part of the related region of a specific class and then predict again to investigate the confidence value change. We use a public GitHub code to measure the confidence increase (e.g., positive value is better).

\begin{figure*}[!ht]
    \centering
    \includegraphics[width=0.9\linewidth]{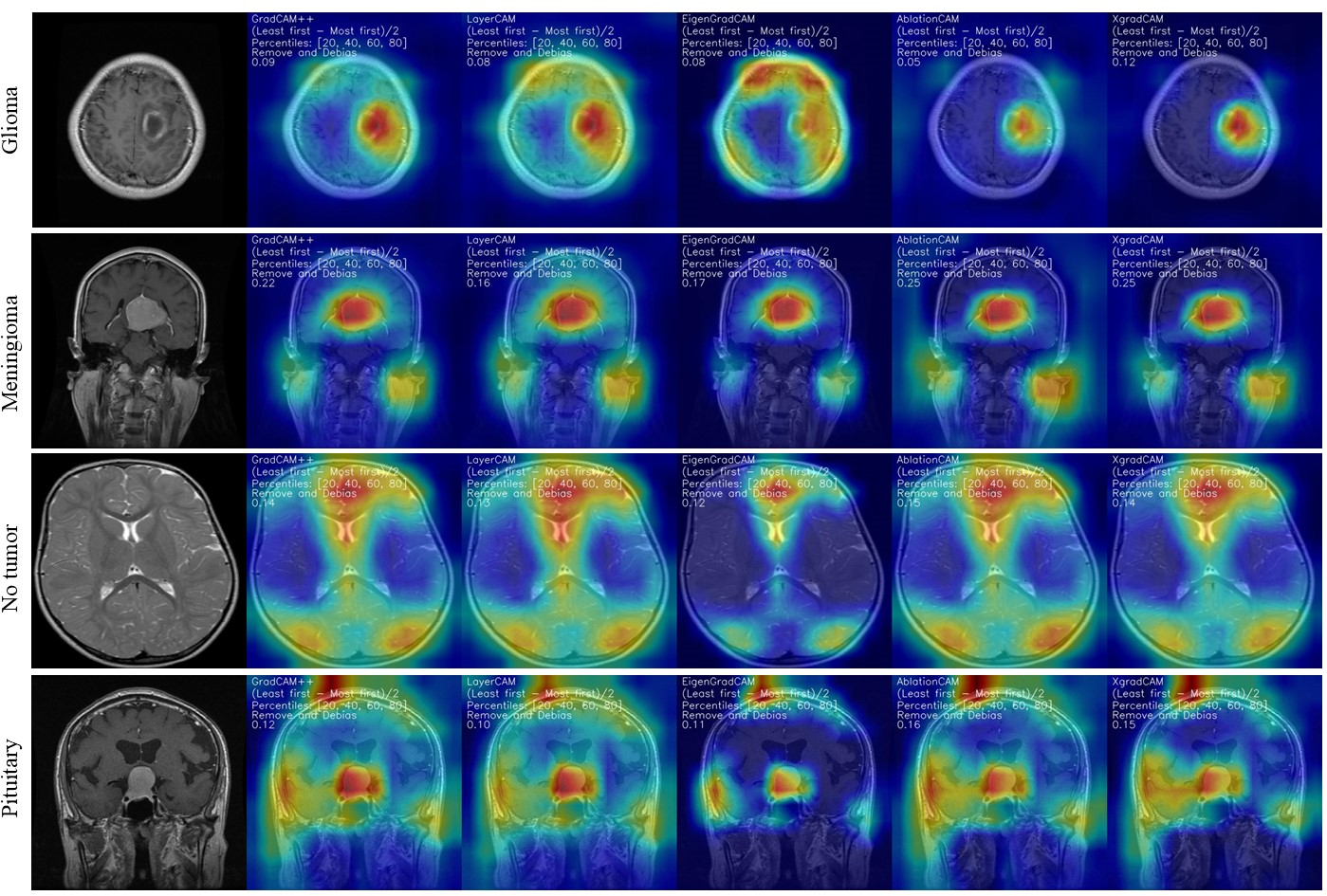}
    \caption{Heatmap visualization using five XAI techniques (ResNet50) and the quantitative metrics using ROAD method in brain tumor dataset. The model attention is more focused on its current location when the color is deeper, such as red. Note that the number as shown in each sub-figure is the averaged confidence increase measured using ROAD technique across four thresholds (positive is better).}
    \label{fig:xai_brain}
\end{figure*}

\begin{figure*}[!ht]
    \centering
    \includegraphics[width=0.9\linewidth]{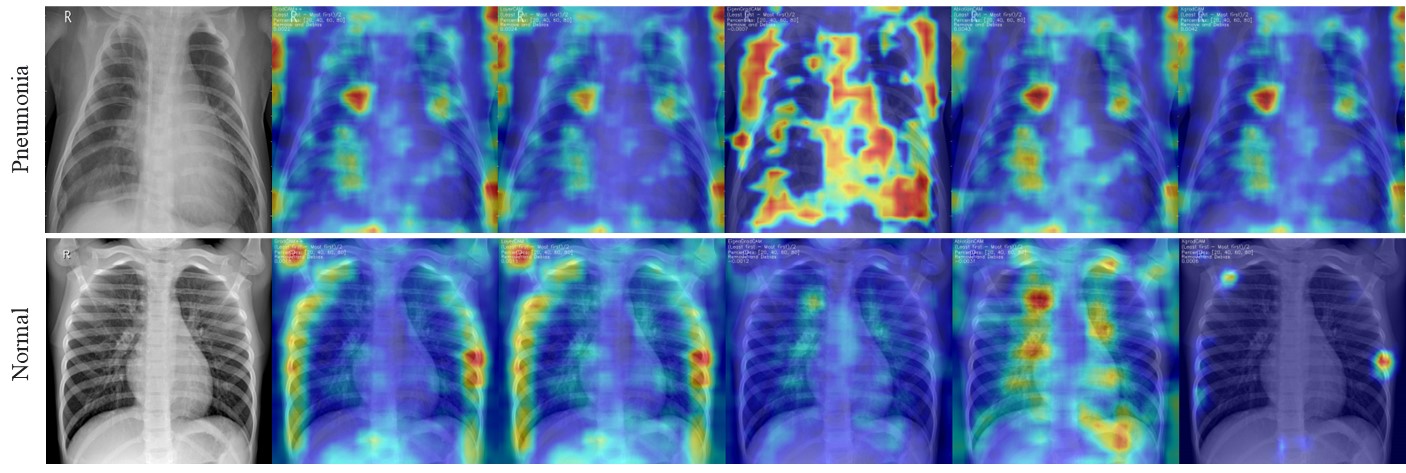}
    \caption{Heatmap visualization using five XAI techniques (ResNet50) and the quantitative metrics using ROAD method in chest x-ray dataset. }
    \label{fig:xai_chest}
\end{figure*}

\begin{figure*}[!ht]
    \centering
    \includegraphics[width=0.9\linewidth]{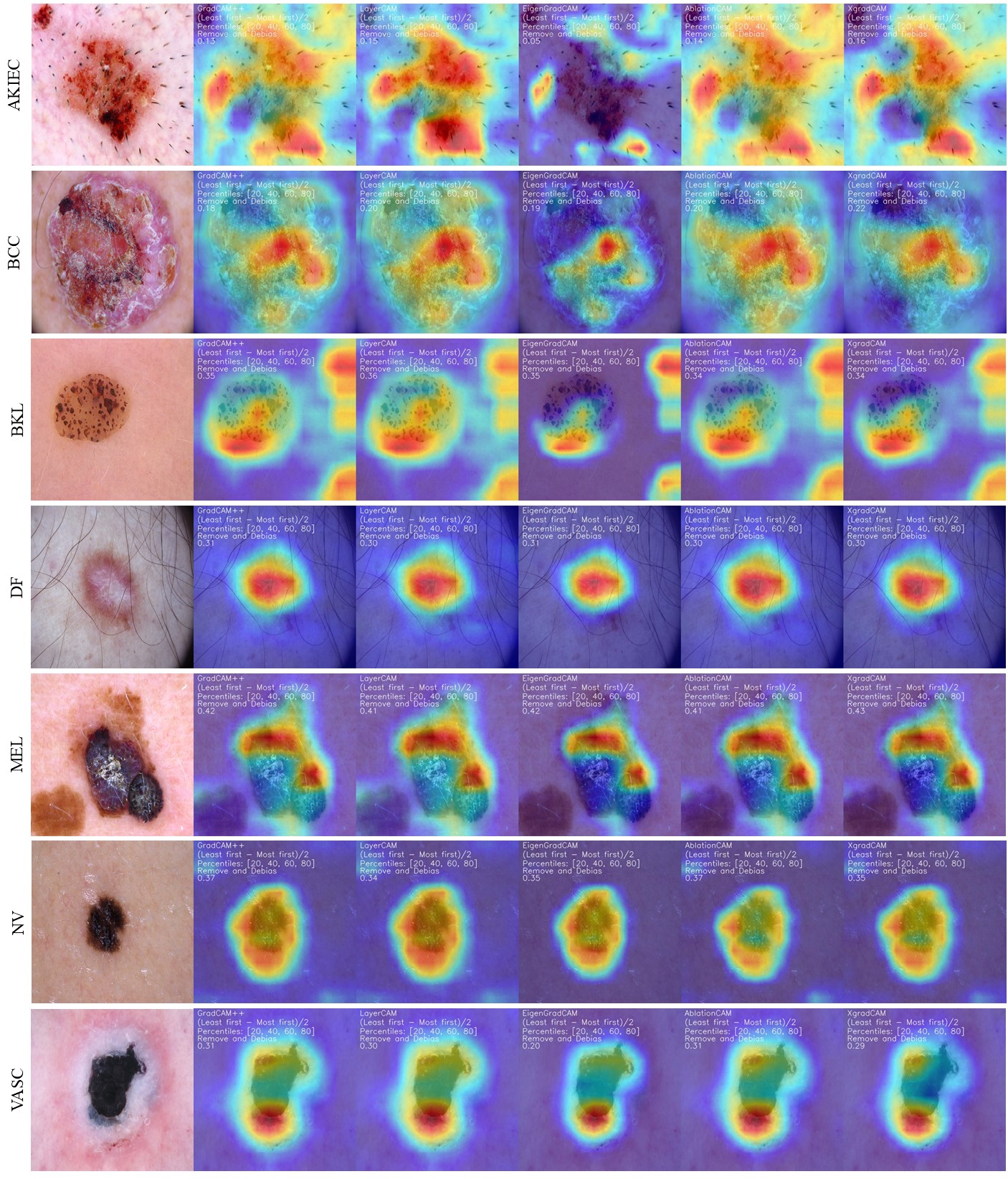}
    \caption{Heatmap visualization using five XAI techniques (ResNet50) and the quantitative metrics using ROAD method in skin cancer dataset.}
    \label{fig:xai_skin}
\end{figure*}

For example, Figure \ref{fig:xai_brain}, Figure \ref{fig:xai_chest}, and Figure \ref{fig:xai_skin} show the explainability analysis with quantitative metrics using five XAI methods. We observed the following points: 1) For brain tumor images, XgradCAM and AblationCAM can identify the abnormal region properly compared to GradCAM++, especially for glioma and meningioma tumors. Also, it demonstrates the highest confidence increase (0.12). This suggests that XgradCAM can provide rich information about increasing the transparency of brain tumor diagnosis. 2) For skin cancer classification, the use of all XAI techniques exhibits remarkable explainability meaning for AKIEC, BCC, DF, NV and VASC except EigenGradCAM, which is less effective for AKIEC and BCC classes. These findings are consistent with its low confidence increase (0.05 in AKIEC). However, for specific classes such as MEL, those XAI approaches can not highlight most of the parts related to abnormal regions.  3) for the chest x-ray dataset, GradCAM++, LayerCAM, AblationCAM and XgradCAM can highlight the abnormal region properly, while EigenGradCAM shows diffused attention on the image. Similarly, AblationCAM and XgradCAM indicate higher confidence increases (0.0043 and 0.0042) compared to EigenGradCAM (-0.0007). This exhibits the potential of AblationCAM and XgradCAM for chest x-ray image diagnosis. In summary, using XAI allows the model to offer feasible transparency, which can assist physicians and patients.

\begin{table}[]
    \centering
    \caption{Executed time (s) of ResNet50 with five XAI techniques in medical datasets. }
    \includegraphics[width=0.97\linewidth]{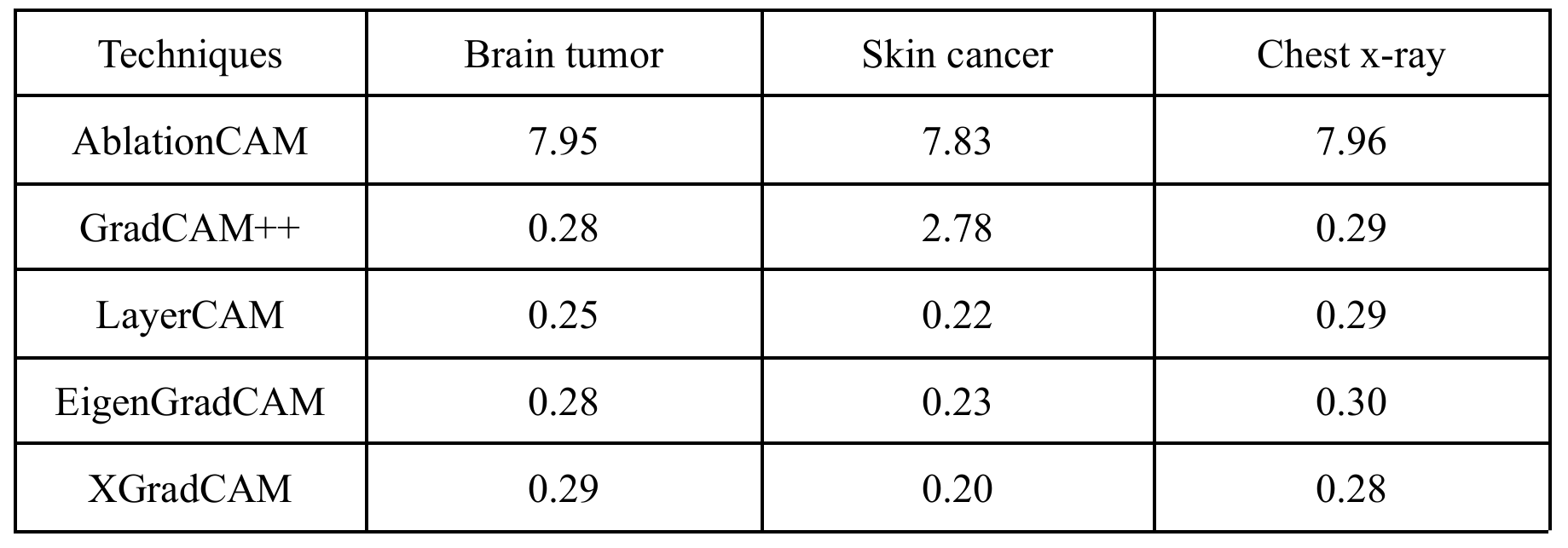}
    
    \label{tab:time}
\end{table}

Furthermore, Table \ref{tab:time} lists the execution time (s) of ResNet50 with five XAI methods in brain tumor, skin cancer, and chest x-ray datasets. As illustrated, the use of LayerCAM, EigenGradCAM and XGradCAM demonstrates feasible time costs, while using AblationCAM and GradCAM++ shows considerable computational burdens. This highlights the potential of LayerCAM in terms of transparency and efficiency.

\begin{table}[]
    \centering
    \caption{Testing classification metrics (\%) using ResNet50 model with public medical data sets.
}
\includegraphics[width=0.97\linewidth]{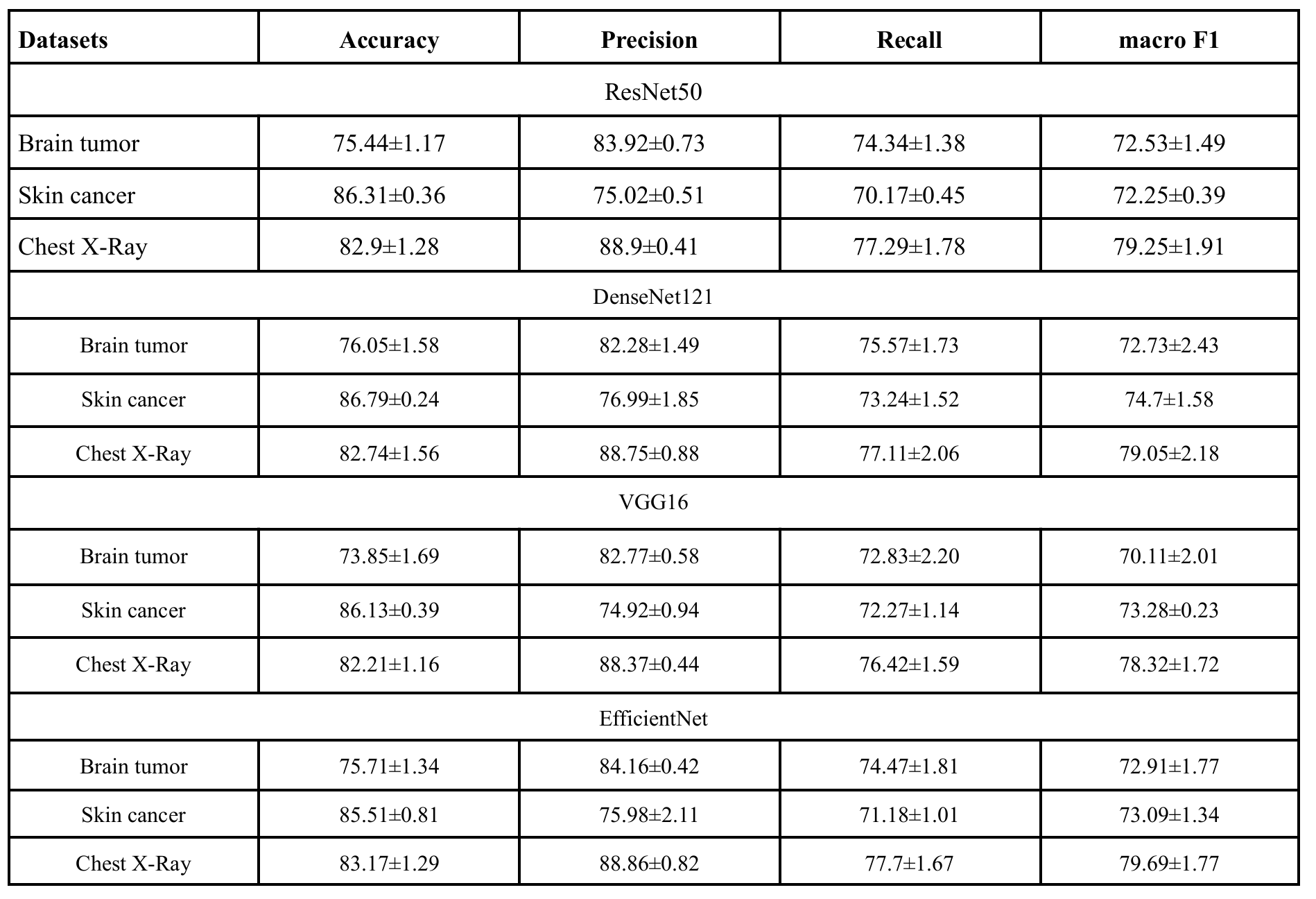}
    
    \label{tab:2}
\end{table}

\section{Discussion}

In general, the results show that XGradCAM excels in visualizing key features of medical images, especially in brain tumor and skin cancer tasks. In contrast, EigenGradCAM has limitations in dealing with complex skin cancer cases. In addition, AblationCAM and GradCAM++ have high computational costs, which limits their wide applicability in real-time clinical environments. LayerCAM, EigenGradCAM, and XGradCAM are more efficient and suitable for resource-constrained clinical environments. XGradCAM combines high efficiency with high explainability, and has great potential for clinical applications.

Specifically, on the brain tumor dataset, XGradCAM and AblationCAM can effectively identify abnormal areas, especially in the classification of gliomas and meningiomas. XGradCAM performs the best, with a confidence gain of 0.12. In the skin cancer dataset, EigenGradCAM showed weaker performance in AKIEC and BCC classes, with a confidence gain of only 0.05. The experimental results of chest X-ray images showed that AblationCAM and XGradCAM were able to better highlight disease-related areas, while EigenGradCAM showed dispersion in this task.

These results could be related to the following reasons:

1) XGradCAM and AblationCAM can effectively locate key regions in images, especially in complex datasets such as brain tumors and chest X-ray images. These methods then can capture local features related to lesions, thus exhibiting higher confidence gains in these tasks.

2) The characteristics of different datasets also have a significant impact on the performance of XAI methods. For example, the image heterogeneity in the skin cancer dataset is high, which makes it challenging for techniques such as EigenGradCAM to effectively capture subtle features when dealing with complex cases. LayerCAM and XGradCAM are better to be managed with this complexity, exhibiting higher stability.

Although our results reveal the strengths and weaknesses of XAI methods, further research is needed. First, hybrid XAI techniques may be developed, such as combining the meticulous focus of LayerCAM with the extensive class-level insights of GradCAM++ to achieve more comprehensive explainability. Second, scalability and efficiency are key challenges for XAI in clinical applications, and high-cost techniques such as AblationCAM need to be optimized, or lightweight XAI methods need to be explored to address these challenges.

Additionally, improving the reliability and consistency of XAI techniques requires the development of higher quality and more detailed visualization tools \cite{wyatt2024explainable}. These tools will facilitate multidimensional data visualization, displaying model feature importance, model uncertainty, and prediction confidence simultaneously, allowing users to interact and analyze model interpretations more easily through high-quality images.

Finally, the versatility of XAI technique needs to be considered with various datasets and clinical tasks. Future research may explore the adaptability of XAI technique in the medical field and multiple imaging modalities to ensure the high explainability of the model.

\section{Conclusion}

This paper experimentally demonstrated the application of XAI techniques in medical image classification and verified the potential of these techniques in improving model transparency and diagnostic accuracy. However, these techniques are not fully explored. First, the experiments are based on only three public medical datasets, and the diversity and scale of the datasets are limited, which may affect the results. Second, the interpretation performance of XAI techniques in complex cases is weak. In addition, this study mainly focuses on image classification tasks, and can be extended to other medical fields in the future, such as sequence data analysis or multimodal data processing, to comprehensively verify the applicability of XAI techniques. In summary, although this paper provides new insights into the application of XAI techniques, further exploration is still needed in terms of dataset diversity, method robustness, and cross-domain application.

\section*{Declaration of competing interest}
The authors declare that they have no known competing
financial interests or personal relationships that could
have appeared to influence the work reported in this
paper.
\section*{Acknowledgment}
This research was supported in part by the National Natural Science Foundation of China (\#82260360), the Guilin Innovation Platform and Talent Program (\#20222C264164), and the Guangxi Science and Technology Base and Talent Project (\#2022AC18004, and \#2022AC21040).

\bibliographystyle{ieeetr}
\bibliography{ref}

\end{document}